\documentclass[fleqn,10pt]{wlpeerj}

\usepackage{amssymb}
\usepackage{latexsym}
\usepackage{amsmath}
\usepackage{subcaption}
\usepackage{url}
\usepackage{hyperref}

\usepackage{xcolor}
\definecolor{newcolor}{rgb}{.8,.349,.1}

\title{Video captioning with stacked attention and semantic hard pull}
\author[1]{Md. Mushfiqur Rahman}
\author[1]{Thasin Abedin}
\author[1]{Khondokar S. S. Prottoy}
\author[1]{Ayana Moshruba}  
\author[2]{Fazlul Hasan Siddiqui}
\affil[1]{Islamic University of Technology, Gazipur, Bangladesh}
\affil[2]{Dhaka University of Engineering and Technology, Gazipur, Bangladesh}
\corrauthor[1]{Md. Mushfiqur Rahman}{mushfiqur11@iut-dhaka.edu}

\begin{abstract}
Video captioning, i.e. the task of generating captions from video sequences creates a bridge between the Natural Language Processing and Computer Vision domains of computer science. The task of generating a semantically accurate description of a video is quite complex. Considering the complexity, of the problem, the results obtained in recent research works are praiseworthy. However, there is plenty of scope for further investigation. This paper addresses this horizon and proposes a novel solution. Most video captioning models comprise two sequential/recurrent layers - one as a video-to-context encoder and the other as a context-to-caption decoder. This paper proposes a novel architecture, namely Semantically Sensible Video Captioning (SSVC) which modifies the context generation mechanism by using two novel approaches - ``stacked attention" and ``spatial hard pull". As there are no exclusive metrics for evaluating video captioning models, we emphasize both quantitative and qualitative analysis of our model. Hence, we have used the BLEU scoring metric for quantitative analysis and have proposed a human evaluation metric for qualitative analysis, namely the Semantic Sensibility (SS) scoring metric. SS Score overcomes the shortcomings of common automated scoring metrics. This paper reports that the use of the aforementioned novelties improves the performance of state-of-the-art architectures.
\end{abstract}

\begin{document}

% \keywords{Video Captioning, Stacked Attention, Spatial Hard Pull, Sequence to Sequence, LSTM}

\flushbottom
\maketitle
\thispagestyle{empty}

\section*{Introduction}

\begin{figure}
     \centering
    {
    \includegraphics[width=.7\linewidth]{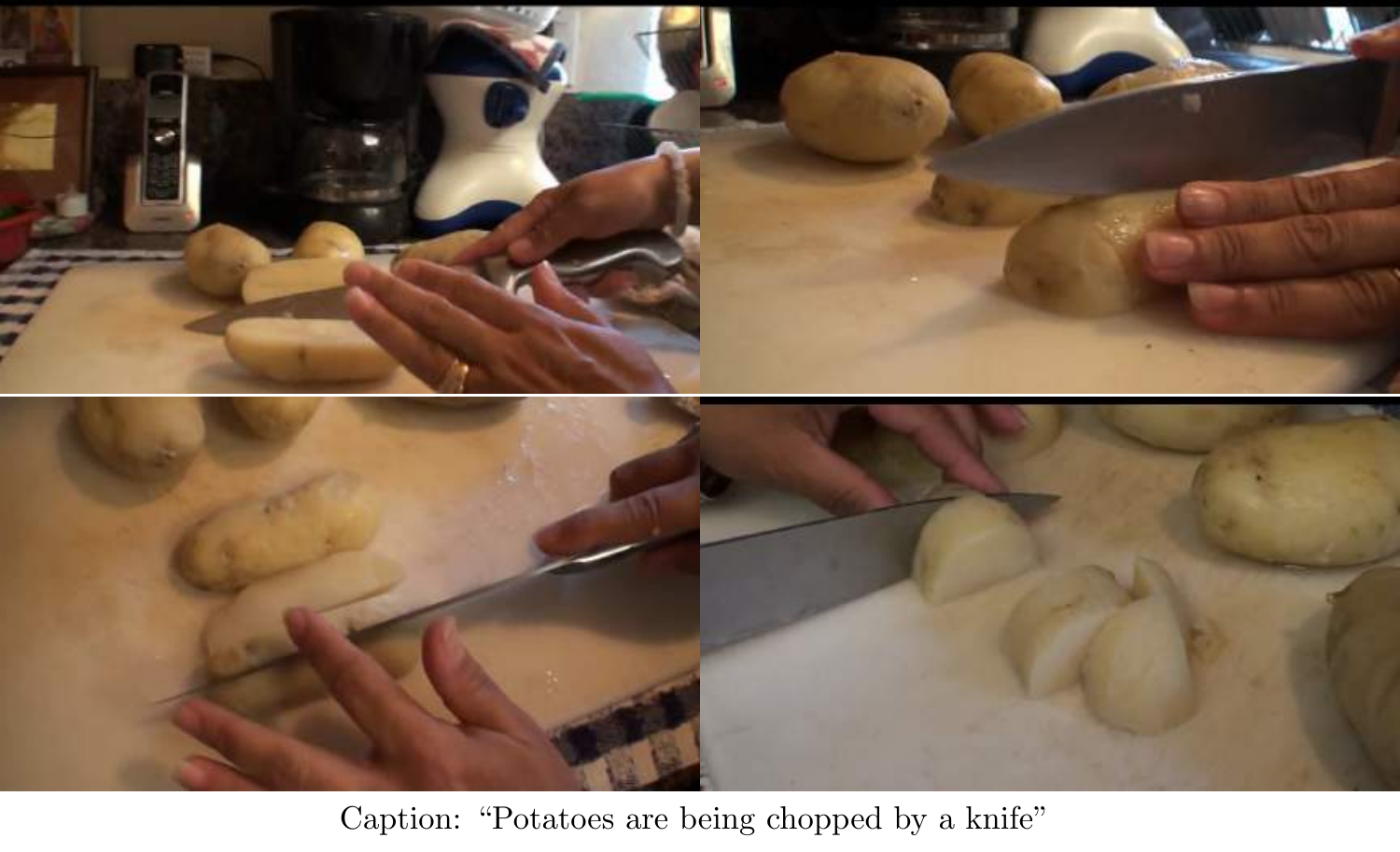}
    %\\caption:“Potatoes are being chopped by a knife”
    }
    \caption{Video Captioning Task}
    \label{fig_video_caption}
\end{figure}

The incredible success in the Image Captioning domain has led the researchers to explore similar avenues like Video Captioning. Video Captioning is the process of describing a video with a complete and coherent caption using Natural Language Processing. The core mechanism of Video Captioning is based on the sequence-to-sequence architecture \citep{gers1999learning}. In video captioning models the encoder encodes the visual stream  and the decoder generates the caption. Such models are capable of retaining both the spatial and temporal information which is essential for generating semantically correct video captions. This requires the video to be split up into a sequence of frames. The model uses these frames as input and generates a series of meaningful words in the form of a caption as output. In Figure ~\ref{fig_video_caption}, an example of a video captioning task has been shown.

Video captioning has many applications, for example, the interaction between humans and machines, aid for people with visual impairments, video indexing, information retrieval, fast video retrieval, etc. Unlike image captioning where only spatial information is required to generate captions, video captioning requires the use of a mechanism that combines spatial information with temporal information to store both the higher level and the lower level features to generate semantically sensible captions. 
Although there are good works in this field, there is still plenty of opportunity for investigation. One of the main opportunities is improving the ability of models to extract high-level features from videos to generate a more meaningful caption. This paper primarily focuses on this aspect.
 
In our paper, we propose a novel architecture that is based on the seq2seq model proposed by \cite{venugopalan2015sequence}. Our novel architecture tries to improve upon this work following the guidelines laid out by preceding literature. Our novel architecture aims to show a possible direction for potential future research.
The goal of our model is to encode a video (presented in the form of a sequence of images) in order to extract information from it and decode encoded data to generate a sentence (presented in the form of a sequence of words). On the encoder side, along with the bi-directional LSTM layers, our model uses the combination of two novel methods - a variation of dual-attention \citep{nam2017dual}, namely, Stacked Attention, and a novel information extraction method, namely, Spatial Hard Pull. The Stacked Attention network sets the priority to the object in the video layer-by-layer. To overcome the redundancy of similar information being lost in the LSTM layers we introduce the Spatial Hard Pull layer. On the decoding side, we employ a sequential decoder with a single layer LSTM and a fully connected layer to generate a word from a given context produced by the encoder.

Most text generation architectures use BLEU \citep{papineni2002bleu} as the scoring metrics. But due to it's inability of considering recall, few variations, including ROUGE \citep{lin2004rouge}, METEOR \citep{banerjee2005meteor} etc., are introduced. Though these automatic scoring metrics are modified in different ways to give more meaningful results, they have their shortcomings \citep{kilickaya2016re,aafaq2019video}. On top of that, no scoring metrics solely used for the purpose of video captioning are available to the best of our knowledge. Some relevant works \citep{xu2017learning,pei2019memory} have used human evaluation. To get a better understanding of the captioning capability of our model, we perform qualitative analysis based on human evaluation and propose our metric, ``Semantic Sensibility Score" or ``SS Score", in short, for video captioning.

\section*{Related Works}
For the past few decades, much work has been conducted on analysing videos to extract different forms of information, such as, sports-feature summary \citep{shih2017survey,ekin2003automatic,ekin2003generic,li2001event}, medical video analysis \citep{quellec2017multiple}, video finger-print \citep{oostveen2002feature} and other high-level features \citep{chang2005columbia,divakaran2003methods,kantorov2014efficient}. These high-level feature extraction mechanisms heavily relied on analyzing each frame separately and therefore, could not retain the sequential information. When the use of memory retaining cells like LSTM \citep{gers1999learning} became computationally possible, models were only then capable of storing meaningful temporal information for complex tasks like caption generation \citep{venugopalan2015sequence}. Previously, caption generation was mostly treated with template based learning approaches \citep{kojima2002natural,xu2015jointly} or other adaptations of statistical machine translation approach \citep{rohrbach2013translating}.

\subsection*{Sequence-to-sequence architecture for video captioning}
Video is a sequence of frames and the output of a video captioning model is a sequence of words. So, video captioning can be classified as a sequence-to-sequence (seq2seq) task. \cite{sutskever2014sequence} introduce the seq2seq architecture where the encoder encodes an input sentence, and the decoder generates a translated sentence. After the remarkable result of seq2seq architecture in different seq2seq tasks \citep{shao2017generating,weiss2017sequence}, it is only intuitive to leverage this architecture in video captioning works like \citep{venugopalan2015sequence}. In recent years, different variations of the base seq2seq architecture has been widely used, e.g. hierarchical approaches \citep{baraldi2017hierarchical,wang2018reconstruction,shih2017survey}, variations of GAN \citep{yang2018video}, boundary-aware encoder approaches \citep{shih2017survey,baraldi2017hierarchical} etc.

\subsection*{Attention in sequence-to-sequence tasks}
In earlier seq2seq literature \citep{venugopalan2015sequence,pan2017video,sutskever2014sequence}, the decoder cells generate the next word from the context of the preceding word and the fixed output of the encoder. As a result, the overall context of the encoded information often got lost and the generated output became highly dependent on the last hidden cell-state.  The introduction of attention mechanism \citep{vaswani2017attention} paved the way to solve this problem. The attention mechanism enables the model to store the context from the start to the end of the sequence. This allows the model to focus on certain input sequences on each stage of output sequence generation \citep{bahdanau2014neural,luong2015effective}.  \cite{luong2015effective} proposed a combined global-local attention mechanism for translation models. In global attention scheme, the whole input is given attention at a time, while the local attention scheme attends to a part of the input at a given time. The work on video captioning enhanced with these ideas. \cite{bin2018describing} describe a bidirectional LSTM model with attention for producing better global contextual representation as well as enhancing the longevity of all the contexts to be recognized. \cite{gao2017video} build a hierarchical decoder with a fused GRU. Their network combines a semantic information based hierarchical GRU, and a semantic-temporal attention based GRU and a multi-modal decoder. \cite{DBLP:journals/corr/BallasYPC15} proposed to leverage the frame spatial topology by introducing an approach to learn spatio-temporal features in videos from intermediate visual representations using GRUs.
 Similarly, several other variations of the attention exists including multi-faceted attention \citep{long2018video}, multi-context fusion attention \citep{wang2018reconstruction} etc. All these papers use one attention at a time. This limits the available information for the respective models. \cite{nam2017dual} introduce a mechanism to use multiple attentions. With their dual attention mechanism, they have retained visual and textual information simultaneously. Ziqi Zhang and team have achieved commendatory scores by  proposing an object relational graph (ORG) based encoder capturing more detailed interaction features and designed a teacher-recommended learning (TRL) method to integrate the abundant linguistic knowledge into the caption model \citep{9156538}.

\section*{Methodology}

As shown in Figure ~\ref{fig_model}, this paper proposes a novel architecture  that uses a combination of stacked-attention (see Figure ~\ref{fig_attn})  and spatial-hard-pull on top of a base video-to-text architecture to generate captions from video sequences. This paper refers to this architecture as Semantically Sensible Video Captioning (SSVC).

\subsection*{Data Pre-processing and Representation}
The primary input of the model is a video sequence. The data pre-processor converts a video clip into a usable video sequence format of 15 frames before passing it to the actual model. Each converted video sequence contains 15 frames placed separated by an equal time gap. The primary output of the model is a sequence of words. The words are stacked to generate the required caption.

\subsubsection*{Visual Feature Extraction}
A video is nothing but a sequence of frames. Each frame is a 2D image with n channels. In sequential architectures, either the frames are directly passed into ConvLSTM \citep{xingjian2015convolutional} layer(s) or the frames are individually passed through a convolutional block and then are passed into LSTM \citep{gers1999learning} layer(s). For our computational limitations, our model uses the latter option. Like ``Sequence to Sequence--Video to Text" \citep{venugopalan2015sequence}, our model uses a pre-trained VGG16 model \citep{simonyan2014deep} and extracts the \emph{fc7} layer's output. This CNN layer converts each $(256 \times 256 \times 3)$ shaped frame into $(1 \times 4096)$ shaped vectors. These vectors are primary inputs of our model.

\subsubsection*{Textual Feature Representation}
Each video sequence has multiple corresponding captions and each caption has a variable number of words. In our model, to  create captions of equal length all the captions are padded with "pad" markers. The "pad" markers help create a uniformity in the data structure. The inclusion of "pad" markers do not create any change in the output as they are omitted during the conversion of tokenized words to complete sentences. 
A "start" marker and an "end" marker marks the start and end of each caption. The entire text data is tokenized, and each word is represented by a one-hot vector of shape $(1 \times unique word count)$. So, a caption with $m$ words is represented with a matrix of shape $(m \times 1 \times unique word count)$. Instead of using these one-hot vectors directly, our model embeds each word into vectors of shape $(1 \times embedding dimension)$ with a pre-trained embedding layer. The embedded vectors are semantically different and linearly distant in vector space from others on the basis of relationship of the corresponding words.

\subsection*{Base architecture}
\begin{figure}[hbt!]
    \centering
    \includegraphics[width=1\textwidth]{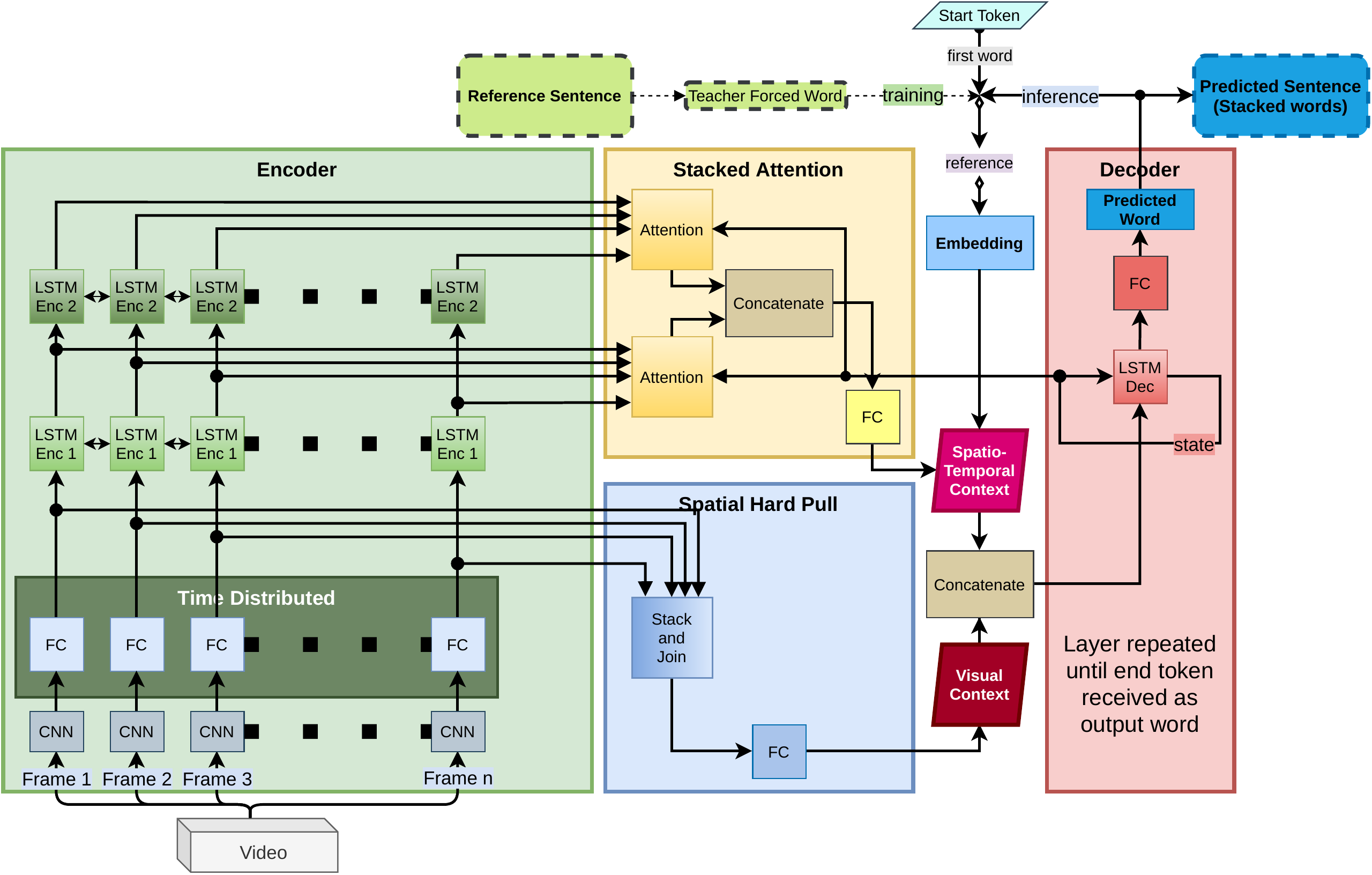}
    \caption{Proposed model with Stacked Attention and Spatial Hard Pull}
    \label{fig_model}
\end{figure}

Like most sequence-to-sequence models, our base architecture consists of a sequential encoder and a sequential decoder. The encoder converts the sequential input vectors into contexts and the decoder converts those contexts into captions. This work proposes an encoder with double LSTM layers with stacked attention. The introduction of a mechanism that stacks attention and the mechanism of pulling spatial information from input vectors are the two novel concepts in this paper and are discussed in detail in later sections. The purpose of using the hard-pull layer is to bring superior extraction capabilities to the model. Since the rest of the model relies on time-series information, the hard-pull layer is necessary for combining information from separate frames and extract general information. The purpose of stacking attention layers is to attain a higher quality temporal information retrieval capability.

\subsubsection*{Multi-layered Sequential Encoder}
The proposed method uses a time-distributed fully connected layer followed by two consecutive bi-directional LSTM layers. The fully connected layer works on each frame separately and then their output moves to the LSTM layers. In sequence-to-sequence literature, it is common to use stacked LSTM for encoder. For it, our intuition is, the two layers capture separate information from the video sequence. Figure ~\ref{fig_stacked_result} show having two layers ensures optimum performance. The output of the encoder is converted into a context. In relevant literature, this context is mostly generated using a single attention layer. This is where this paper proposes a novel concept. With the mechanism mentioned in later sections our model generates a spatio-temporal context.

\subsubsection*{Single-layered Sequential Decoder}
The proposed decoder uses a single layer LSTM followed by a fully connected layer to generate a word from a given context. In relevant literature, many models have used stacked decoder. Most of these papers suggest, each layer of decoder handles separate information, while our model uses a single layer. Our experimental results show that having stacked decoder does not improve the result much for our architecture. Therefore, instead of stacking decoder layers, we increased the number of decoder cells. Specifically, we have used twice as many cells in decoder than in encoder and it has shown the optimum output during experimentation.

\subsubsection*{Training and Inference Behaviour}
To mark the start of a caption and to distinguish the mark from the real caption, a ``start" token is used at the beginning. The decoder uses this token as a reference to generate the first true word of the caption. Figure ~\ref{fig_model} represents this as ``first word".
During inference, each subsequent word is generated with the previously generated word as reference. The sequentially generated words together form the desired caption. The loop terminates upon receiving the ``end" marker. 

During training, if each iteration in the generation loop uses previously generated word, then one wrong generation can derail the entire remaining caption. Thus error calculation process becomes vulnerable. To overcome this, like most seq-to-seq papers, we use the teacher forcing mechanism \citep{lamb2016professor}. The method uses words from the original caption as reference for generating the next words during the training loop. Therefore, the generation of each word is independent of previously generated words. Figure ~\ref{fig_model} illustrates this difference in training and testing time behaviour. During training, ``Teacher Forced Word" is the word from the reference caption for that iteration. 

\subsection*{Proposed Context Generation architecture}
\label{proposed}
The paper proposes two novel methods. The methods show promising signs to make progress in the field of video captioning.

\subsubsection*{Stacked Attention}

\begin{figure}
     \centering
    {
    \includegraphics[width=\linewidth]{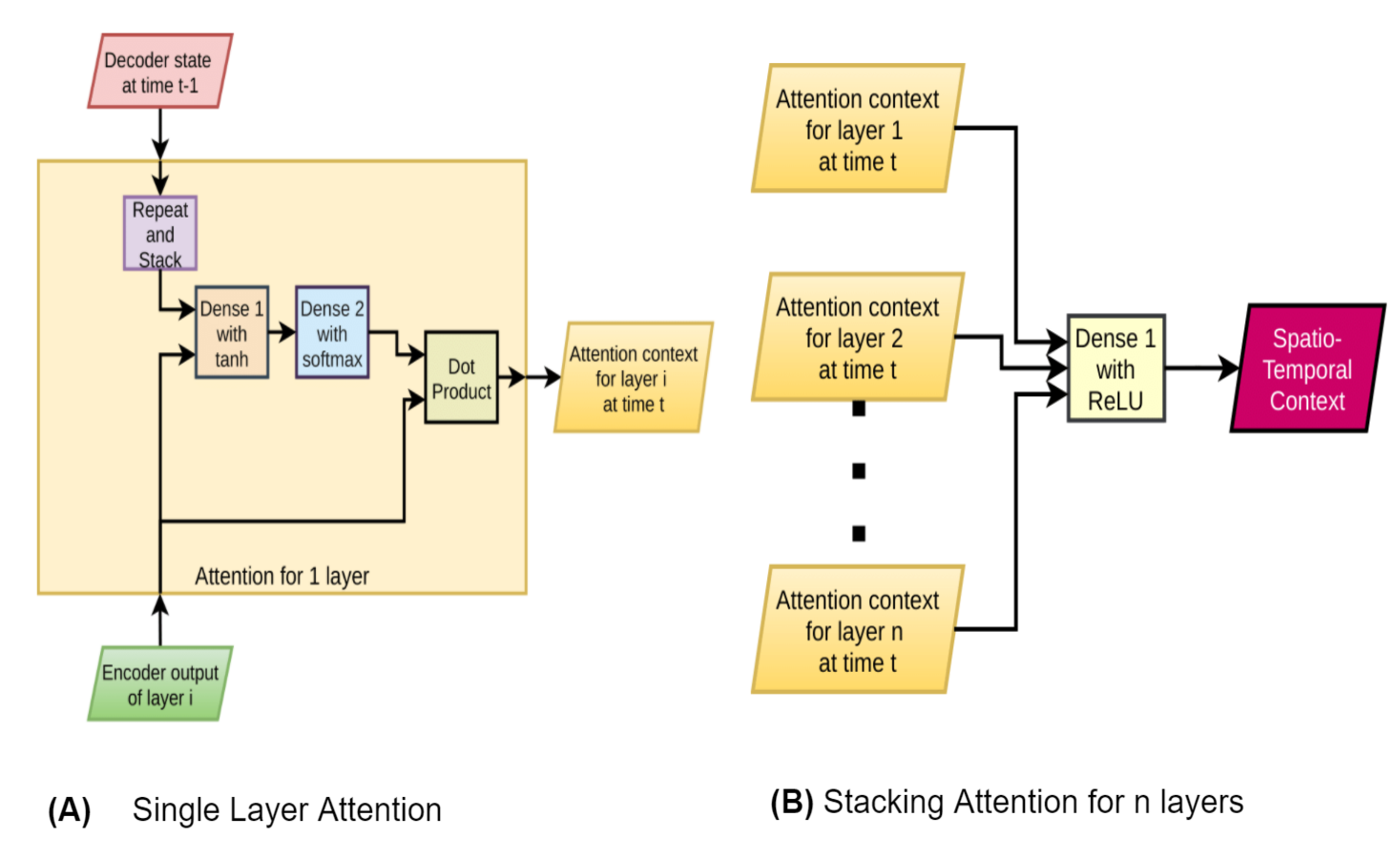}
    }
    \caption{Diagram of Stacked Attention}
    \label{fig_attn}
\end{figure}

Attention creates an importance map for individual vectors from a sequence of vectors. In text-to-text, i.e., translation models, this mapping creates a valuable information that suggests which word or phrase in the input side has higher correlation to which words and phrases in the output. However, in video captioning, attention plays a different role. For a particular word, instead of determining which frame (from original video) or frames to put more emphasis on, the stacked attention emphasizes on objects. This paper uses a stacked LSTM. Like other relevant literature \citep{venugopalan2015sequence,song2017hierarchical}, this paper reports separate layers to carry separate information. So, if each layer has separate information, it is only intuitive to generate separate attention for each layer. Our architecture stacks the separately generated attentions and connects them with a fully connected layer with \emph{tanh} activation. The output of this layer determines whether to put more emphasis on the object or the action.

\begin{equation}
f_{attn}(\left[h,ss\right]) = a_{s}(W_{2}* a_{tanh}(W_{1}\left[h,ss\right] + b_{1}) + b_{2})
\label{eq_attn}
\end{equation}
\begin{equation}
c_{attn} = dot(h,f_{attn}(\left[h,ss\right]))
\label{eq_context}
\end{equation}
\begin{equation}
c_{st} = a_{relu}(W_{st}[c_{attn_{1}}, c_{attn_{2}},\dots,c_{attn_{n}}] + b_{st})
\label{eq_stacked_context}
\end{equation}
where,
\begin{itemize}[noitemsep, label={}]
    \item $h=$ encoder output for 1 layer
    \item $ss=$ decoder state is repeated to match $h$'s dimension
    \item $a_{s}(x)= \frac{\exp{(x-\max(x))}}{\sum{\exp{(x-\max(x))atten}}}$
    \item $n=$ number of attention layers to be stacked
    \item $c_{attn}=$ context for single attention
    \item $dot( )$ function represents scalar dot-product 
    \item $c_{st}=$ stacked context for $n$ encoder layers
\end{itemize}

Eq. ~\ref{eq_attn} is the attention function. Eq. ~\ref{eq_context} uses the output of this function to generate the attention context for one layer. Eq. ~\ref{eq_stacked_context} combines the attention context of several layers to generate the desirable spatio-temporal context. The paper also refers to this context as ``stacked context". Figure ~\ref{fig_attn} corresponds with these equation. In SSVC, we have particularly used $n=2$, where $n$ is the number of attention layers in the stacked attention.

The stacked attention mechanism generates the spatio-temporal context for the input video sequence. All types of low-level context required to generate the next word is available in this novel context generation mechanism.

\subsubsection*{Spatial Hard Pull}
\cite{amaresh2019video} mentions that most successful image and video captioning models mainly learn to map low-level visual features to sentences. They do not focus on the high-level semantic video concepts - like actions and objects. By low-level features, they meant object shapes and their existence in the video. High-level features refer to proper object classification with position in the video and the context in which the object appears in the video. On the other hand, our analysis of previous architectures shows that almost identical information is often found in nearby frames of a video. However, passing the frames through LSTM layer does not help to extract any valuable information from this almost identical information. So, we have devised a method to hard-pull the output of the time-distributed layer and use it to add high-level visual information to the context. This method enables us to extract meaningful high-level features, like objects and their relative position in the individual frames.

This method extracts information from all frames simultaneously and does not consider sequential information. As the layer pulls spatial information from sparsely located frames, this paper names it ``Spatial Hard Pull" layer. It can be compared to a skip connection. But unlike other skip connections, it skips a recurrent layer, and directly contributes to the context. The output units of the fully connected (FC) layer of this spatial-hard-pull layers determines how much effect will the sparse layer have on the context. Figure ~\ref{fig_join_result} indicate the performance improvement in the early stages due to SHP layer and the fall of scores in the later stages due to high variance.

\section*{Proposed Scoring Metric}
No automatic scoring metric has been designed yet for the sole purpose of video captioning. The existing metrics that have been built for other purposes, like neural machine translation, image captioning, etc., are used for evaluating video captioning models. For quantitative analysis, we use the BLEU scoring metric \citep{papineni2002bleu}. Although these metrics serve similar purposes, according to  \cite{aafaq2019video}, they fall short in generating ``meaningful" scores for video captioning.

\begin{figure}
     \centering
    {
    \includegraphics[width=.9\linewidth]{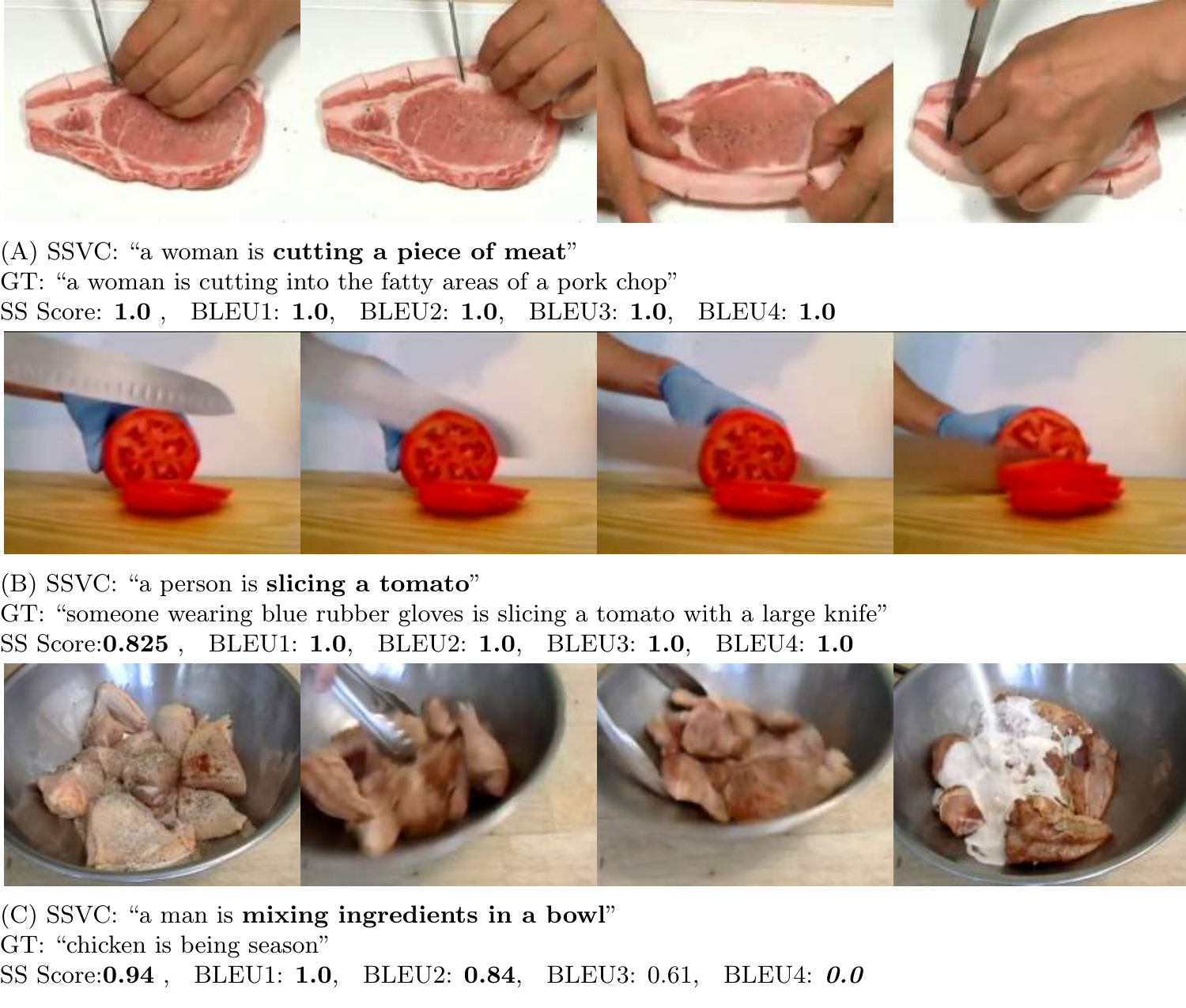}
    }
    \caption{In Figure(6A) [Source: \href{https://www.youtube.com/watch?v=6t0BpjwYKco\&t=230s}{https://www.youtube.com/watch?v=6t0BpjwYKco\&t=230s}] and Figure(6B) [Source: \href{https://www.youtube.com/watch?v=j2Dhf-xFUxU\&t=20s}{https://www.youtube.com/watch?v=j2Dhf-xFUxU\&t=20s}], our model is able to extract the action part correctly and gets decent score in both SS and BLEU score. In Figure(6C) [Source: \href{https://www.youtube.com/watch?v=uxEhH6MPH28\&t=29s}{https://www.youtube.com/watch?v=uxEhH6MPH28\&t=29s}], the output is perfect and SS Score is high. However, BLEU4 is 0}
    \label{fig_output1}
\end{figure}

\begin{figure}
     \centering
    {
    \includegraphics[width=.9\linewidth]{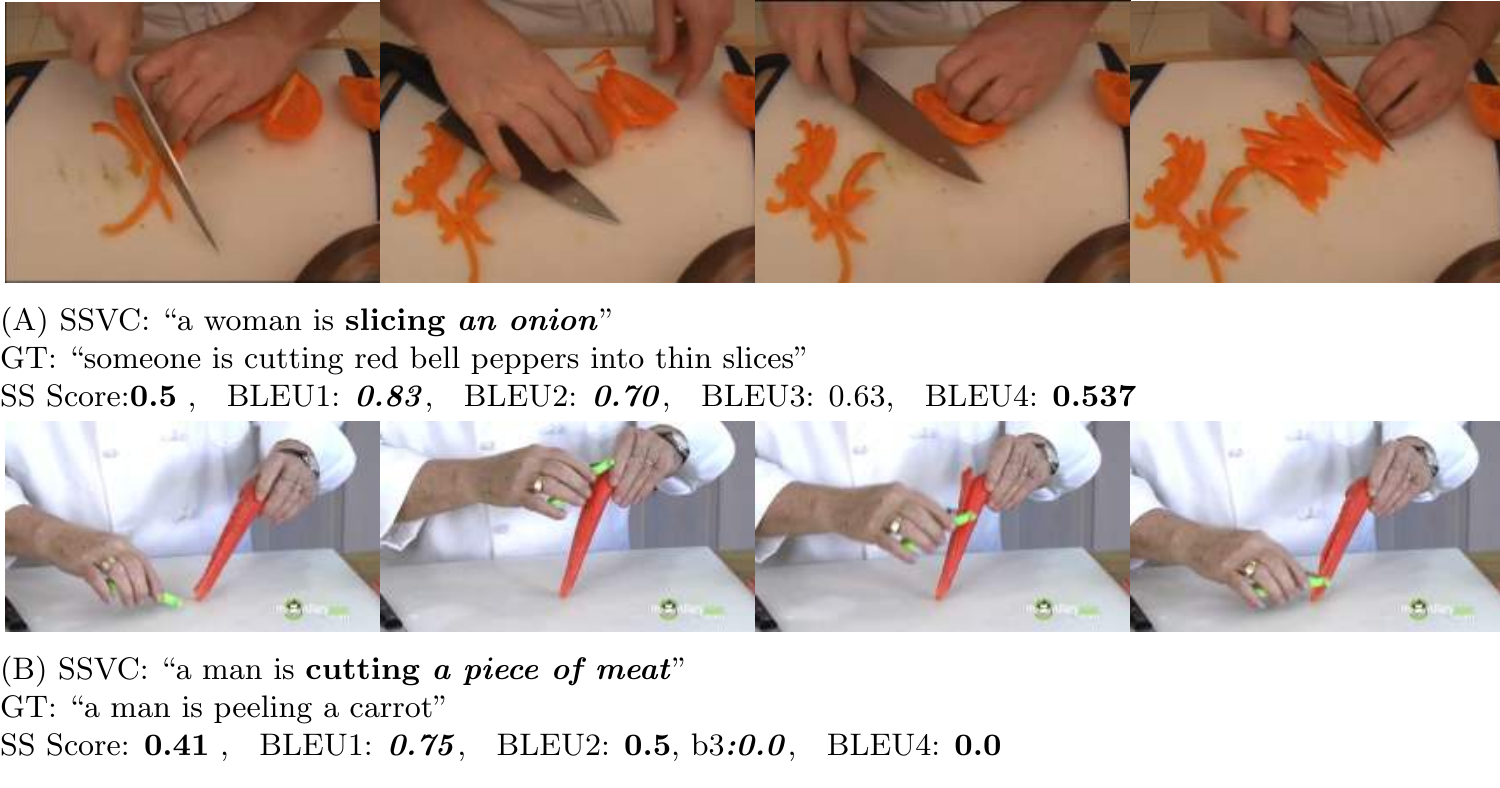}
    }
    \caption{In Figure(7A) [Source: \href{https://www.youtube.com/watch?v=VahnQw2gTQY\&t=298s}{https://www.youtube.com/watch?v=VahnQw2gTQY\&t=298s}] and Figure(7B) [Source: \href{https://www.youtube.com/watch?v=YS1mzzhmWWA\&t=9s}{https://www.youtube.com/watch?v=YS1mzzhmWWA\&t=9s}], our model is able to extract only the action part correctly. The generated caption gets mediocre score in both SS and BLEU score.}
    \label{fig_output2}
\end{figure}

\begin{figure}
     \centering
    {
    \includegraphics[width=.9\linewidth]{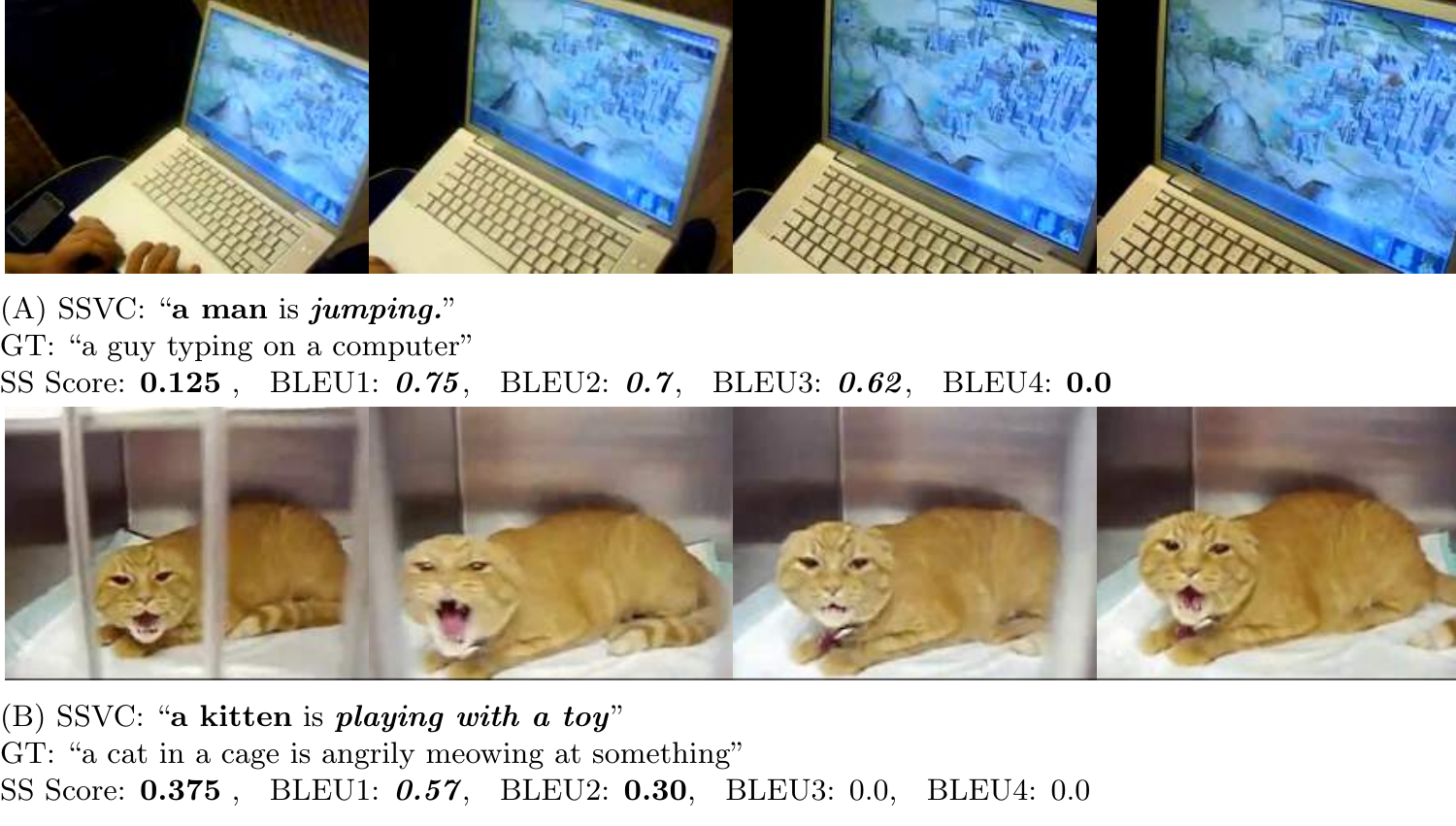}
    }
    \caption{In Figure(8A) [Source: \href{https://www.youtube.com/watch?v=R2DvpPTfl-E\&t=20s}{https://www.youtube.com/watch?v=R2DvpPTfl-E\&t=20s}] and Figure(8B) [Source: \href{https://www.youtube.com/watch?v=1hPxGmTGarM\&t=9s}{https://www.youtube.com/watch?v=1hPxGmTGarM\&t=9s}], the generated caption is completely wrong in case of actions, but BLEU1 gives a very high score. On the contrary, SS Score heavily penalizes them.}
    \label{fig_output3}
\end{figure}

\begin{figure}
     \centering
    {
    \includegraphics[width=.9\linewidth]{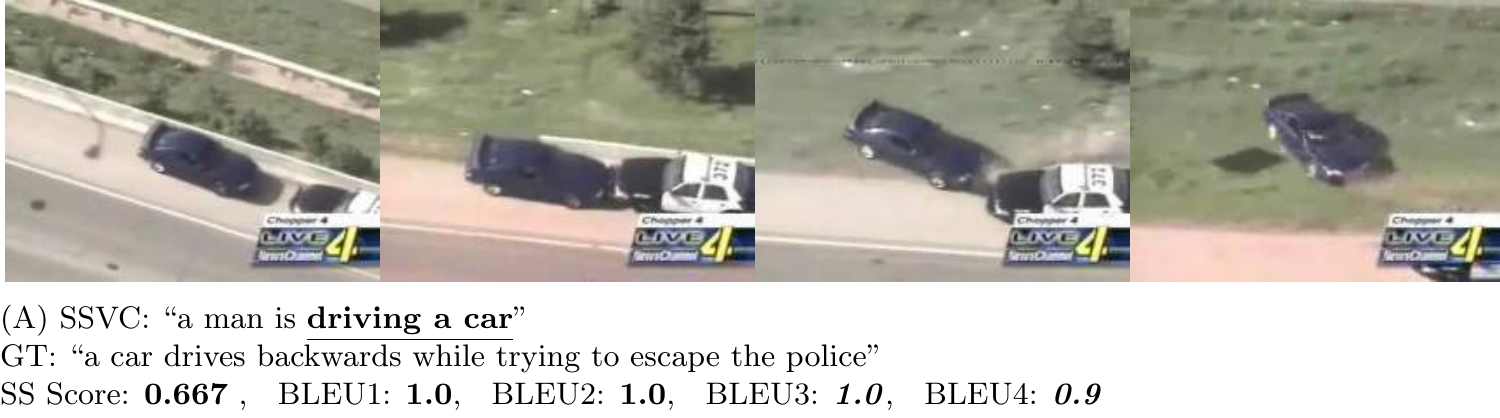}
    }
    \caption{In Figure(9) [Source: \href{https://www.youtube.com/watch?v=3opDcpPxllE\&t=50s}{https://www.youtube.com/watch?v=3opDcpPxllE\&t=50s}] a car is running away and the police chasing it is happening simultaneously. Our SSVC model only predicts the driving part. Thus the generated captions only partially capture the original idea. However, BLEU evaluates them with very high score where SS Score evaluates them accordingly.}
    \label{fig_output4}
\end{figure}

\begin{figure}
     \centering
    {
    \includegraphics[width=.9\linewidth]{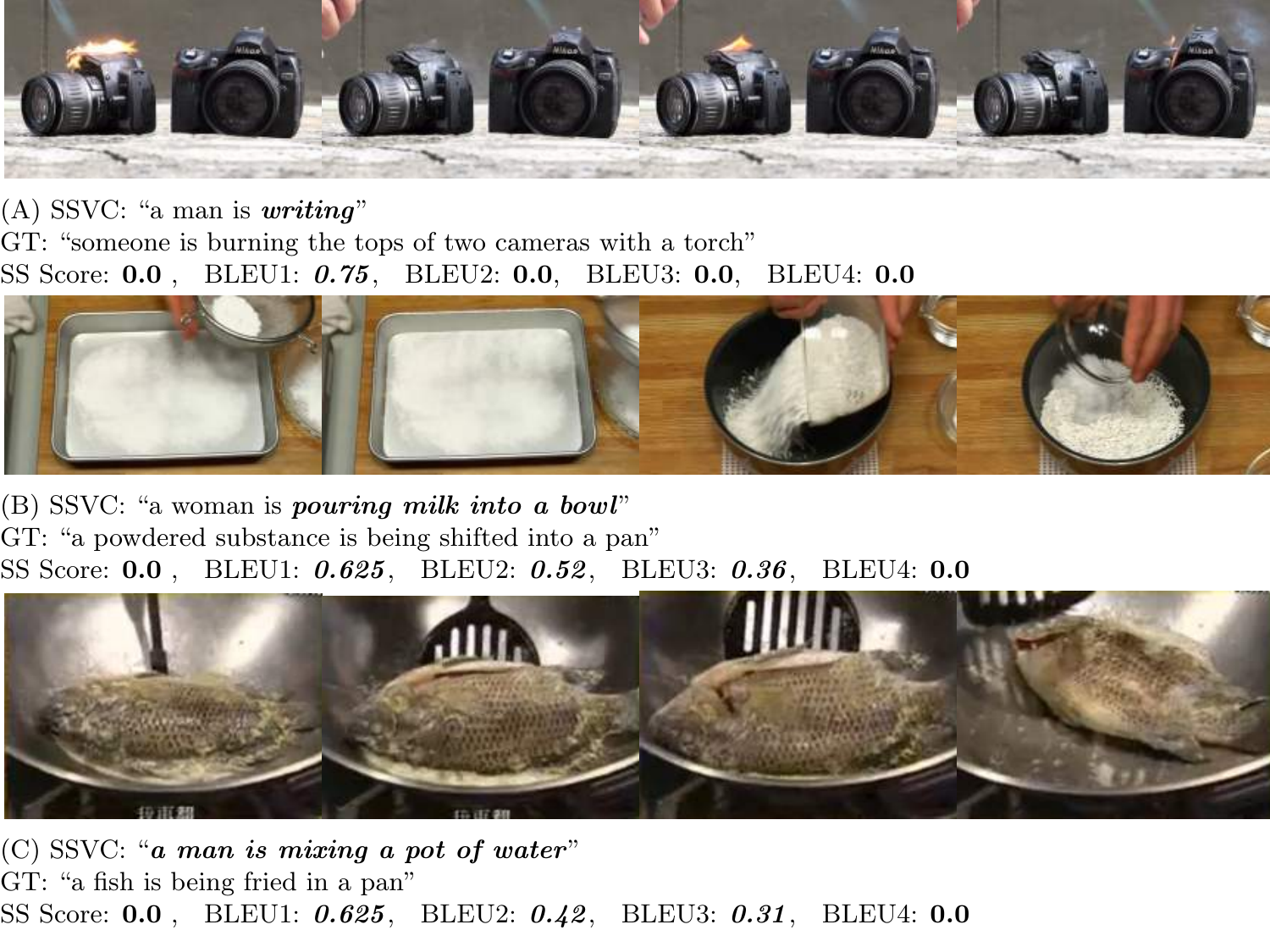}
    }
    \caption{In Figure(10A) [Source: \href{https://www.youtube.com/watch?v=D1tTBncIsm8\&t=841s}{https://www.youtube.com/watch?v=D1tTBncIsm8\&t=841s}], Figure(10B) [Source: \href{https://www.youtube.com/watch?v=Cv5LsqKUXc\&t=71s}{https://www.youtube.com/watch?v=Cv5LsqKUXc\&t=71s}], and Figure(10C) [Source: \href{https://www.youtube.com/watch?v=2FLsMPsywRc\&t=45s}{https://www.youtube.com/watch?v=2FLsMPsywRc\&t=45s}], the generated caption is completely wrong, but BLEU1 gives a very high score where SS Score gives straight up zero. So BLEU performs poorly here.}
    \label{fig_output5}
\end{figure}

BLEU is a precision-based metric. It is mainly designed to evaluate text at a corpus level. BLEU metric can be calculated in reference to 1 sentence or in reference to a corpus of sentences \citep{BleuWeb}. Though the BLEU scoring metric is widely used, \cite{post2018call,callison2006re,graham2015re} demonstrate the inefficiency of BLEU scoring metric in generating a meaningful score for tasks like video captioning. A video may have multiple contexts. So, machines face difficulty to accurately measure the merit of the generated captions as there is no specific right answer. Therefore, for video captioning, it is more challenging to generate meaningful scores to reflect the captioning capability of the model. As a result, human evaluation is an important part to judge the effectiveness of the captioning model. In fact, Figure ~\ref{fig_output1}, ~\ref{fig_output2}, ~\ref{fig_output3}, ~\ref{fig_output4}, and ~\ref{fig_output5} show this same fact that a higher BLEU score is not necessarily a good reflection of the captioning capability. On the other hand, our proposed human evaluation method portrays a better reflection of the model's performance compared to the BLEU scores.

\subsection*{Semantic Sensibility(SS) Score Evaluation}
To get a better understanding of captioning capability of our model, we perform qualitative analysis that is based on human evaluation similar to \citep{graham2018evaluation,xu2017learning,pei2019memory}. We propose a human evaluation metric, namely ``Semantic Sensibility" score, for video captioning. It evaluates sentences at a contextual level from videos based on both recall and precision. It takes 3 factors into consideration. These are the grammatical structure of predicted sentences, detection of the most important element (subject or object) in the videos and whether the captions give an exact or synonymous analogy to the action of the videos to describe the overall context.

It is to be noted that for the latter two factors, we take into consideration both the recall and precision values according to their general definition. In case of recall, we evaluate these 3 factors from our predicted captions and match them with the corresponding video samples. Similarly, for precision, we judge these factors from the video samples and match them with the corresponding predicted captions. Following such comparisons, each variable is assigned to a boolean value of 1 or 0 based on human judgment. The significance of the variables and how to assign their values are elaborated below:

\subsubsection*{$S_{grammar}$}
\begin{equation}
    S_{grammar}= 
        \begin{cases}
            1, & \text{if grammatically correct}\\
            0, & \text{otherwise}
        \end{cases}
    \label{eq_grammar}
\end{equation}
$S_{grammar}$ evaluates the correctness of grammar of the generated caption without considering the video.

\subsubsection*{$S_{element}$}
\begin{equation}
    S_{ element} = \frac{\frac{1}{R}\sum\limits_{i=1}^{R}S_{element_{recall}^{i}} 
    + \frac{1}{P}\sum\limits_{i=1}^{P}S_{element_{precision}^{i}}}{2}
    \label{eq_element}
\end{equation}
where,
\begin{itemize}[noitemsep, label={}]
    \item $R$ = number of prominent objects in video
    \item $P$ = number of prominent objects in caption
\end{itemize}
As $S_{action}$ evaluates the action-similarity between the predicted caption and its corresponding video, $S_{element}$ evaluates the object-similarity. For each object in the caption, the corresponding $S_{element_{precision}}$ receives a boolean score and for the major objects in the video, the corresponding $S_{element_{recall}}$ receives a boolean score. The average recall and average precision is combined to get the $S_{element}$.

\subsubsection*{$S_{action}$}
\begin{equation}
    S_{action} = \frac{S_{action_{recall}} + S_{action_{ precision}}}{2}
    \label{eq_action}
\end{equation}
$S_{action}$ evaluates the ability to describe the action-similarity between the predicted caption and its corresponding video. $S_{action_{recall}}$ and $S_{action_{precision}}$ separately receives a boolean score (1 for correct, 0 for incorrect) for action recall and action precision respectively. By action recall, we determine if the generated caption has successfully captured the most prominent action of the video segment. Similarly, by action precision, we determine if the action mentioned in the generated caption is present in the video or not.

\subsubsection*{SS Score calculation}
Combining equations Eq.~\ref{eq_grammar}, Eq.~\ref{eq_element} and Eq.~\ref{eq_action}, the equation for the SS Score can be obtained.
\begin{equation}
    SS Score = \frac{1}{N}\sum\limits_{n=1}^N\left(S_{ grammar}*\frac{S_{ element} + S_{ action} }{2}\right)
    \label{eq_ss}
\end{equation}

During this research work, the SS Score was calculated by 4 final-year undergraduate students studying at the Department of Computer Science and Engineering at the Islamic University of Technology. They are all Machine Learning researchers and are fluent English speakers. Each caption was separately scored by at least two annotators to make the scoring consistent and accurate.

\section*{Results}

\subsection*{Dataset And Experimental Setup}

Our experiments are primarily centered around comparing our novel model with different commonly used architectures for video captioning like simple attention \citep{gao2017video,wu2018interpretable}, modifications of attention mechanism \citep{yang2018video,yan2019stat,zhang2019show}, variations of visual feature extraction techniques \citep{aafaq2019spatio,wang2018m3} etc that provide state-of-the-art results. We conducted the experiments under identical computational environment - Framework: \emph{Tensorflow 2.0}, Platform: \emph{Google Cloud Platform with a virtual machine having an 8-core processor and 30GB RAM}, GPU: \emph{None}. We used the Microsoft-Research Video Description (MSVD) dataset \citep{chen2011collecting}. It contains 1970 video snippets together with 40 English captions \citep{chen2020delving} for each video. We split the entire dataset into training, validation, and test set with 1200, 100, and 670 snippets respectively following previous works \citep{venugopalan2015sequence,pan2016jointly}. To create a data-sequence, frames from a video are taken with a fixed temporal distance. We used 15 frames for each data-sequence. After creating the data-sequences, we had almost 65000 samples in our dataset. Though there is a large number of samples in the final dataset, the number of distinct trainable videos are only 1200. 1200 videos is not a large enough number. Having a larger dataset would be better for the training.

For the pre-trained embedding layer, we used `glove.6B.100d' \citep{pennington2014glove}. Due to lack of GPU, we used 256 LSTM units in each encoder layer and 512 LSTM units in our decoder network and trained each experimental model for 40 epochs. To analyse the importance of the Spatial Hard Pull layer, we also tuned the Spatial Hard Pull FC units from 0 to 45 and 60 successively.

One of the most prominent benchmarks for video data is TRECVid. Particularly, the 2018 TRECVid challenge \cite{awad2018trecvid} that included video captioning, video information retrival, activity detection, etc could be an excellent benchmark for our work. However, due to our limitations like lack of enough computational resources, rigidity in data pre-processing due to memory limitation and inability to train on a bigger dataset, we could not analyse our novel model with global benchmarks like TRECVid. On top of that, some of the benchmark models use multiple features as input to the model. However, we only use a single 2D based CNN feature as input as we wanted to make an extensive study on the capability of 2D CNN for video captioning. So, we implemented some of the fundamental concepts used in most state-of-the-art works on our experimental setup with single input 2D CNN feature. Thus, we performed ablation study to make a qualitative and quantitative analysis of our model. The performance of our two proposed novelties shows potential for improvement.

We used the BLEU score as one of the two main scoring criteria. To calculate BLEU on a dataset with multiple ground-truth captions, we used the Corpus BLEU calculation method \citep{BleuWeb}. The BLEU scores reported throughout this paper actually indicates the Corpus BLEU Score. Our proposed architecture, SSVC, with 45 hard-pull units and 2 layer stacked attention gives the BLEU score of \emph{``BLEU1": 0.7072, ``BLEU2": 0.5193, ``BLEU3": 0.3961, ``BLEU4": 0.1886} after 40 epochs of training with the best combination of hyper-parameters. For generating the SS Score, we considered the first 50 random videos from the test set. We obtained an SS Score of 0.34 for the SSVC model.

\subsection*{Ablation Study of Stacked Attention}

\begin{figure}
     \centering
    {
    \includegraphics[width=.7\linewidth]{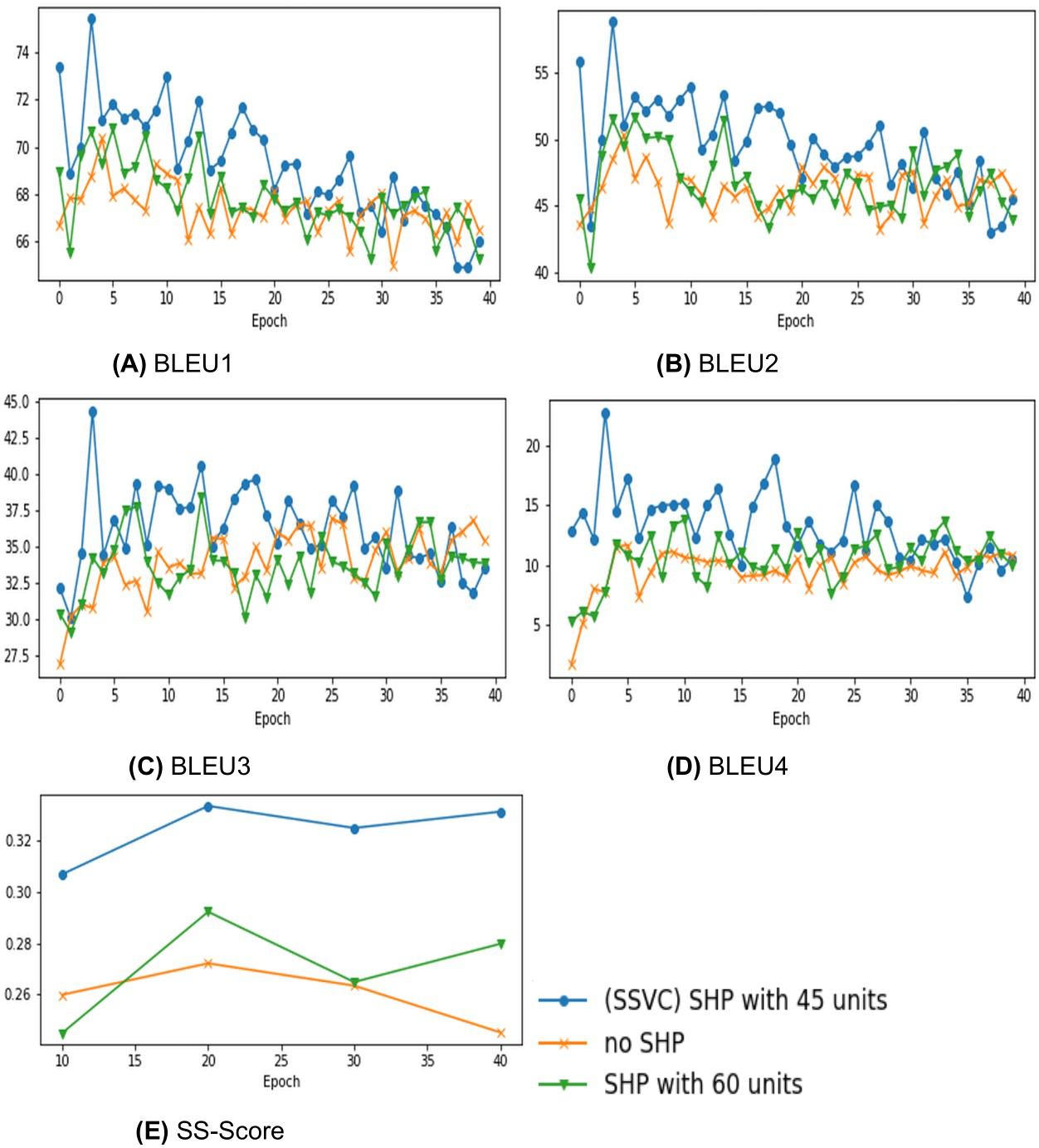}
    %\\caption:“Potatoes are being chopped by a knife”
    }
    \caption{Comparing Stacked Attention with variations in encoder attention architecture}
    \label{fig_stacked_result}
\end{figure}

\begin{itemize}
    \item \textbf{No attention:} Many previous works \citep{long2018video,nam2017dual} mentioned that captioning models perform better with some form of attention mechanism. Thus, in this paper, we avoid comparing use of attention and no attention mechanisms.
    \item \textbf{Non stacked (or single) attention:} In relevant literature, though the use of attention is very common, the use of stacked attention is quite infrequent. \cite{nam2017dual} have shown the use of stacked (or dual) attention and improvements of performance that are possible through it. In Figure ~\ref{fig_stacked_result}, the comparison between single attention and stacked attention indicates dual attention has clear edge over single attention.
    \item \textbf{Triple Attention:} Since the use of dual attention has improved performance in comparison to single attention, it is only evident to create a triple attention to check the performance. Figure ~\ref{fig_stacked_result} show that triple attention under-performs in comparison to all other variants.
\end{itemize}

Considering our limitations, our stacked attention gives satisfactory results for both BLEU and SS Score in comparison to the commonly used attention methods when performed on similar experimental setup. Graphs in Figure ~\ref{fig_stacked_result} suggest the same fact that our stacked attention improves the result of existing methods due to improved overall temporal information. Moreover, we can clearly see that the 2 layer LSTM encoder performs much better than single or triple layer encoder. Combining these two facts, we can conclude that, our dual encoder LSTM with stacked attention has the capability to improve corresponding architectures.

\subsection*{Ablation Study of Spatial Hard Pull}

\begin{figure}
     \centering
    {
    \includegraphics[width=.7\linewidth]{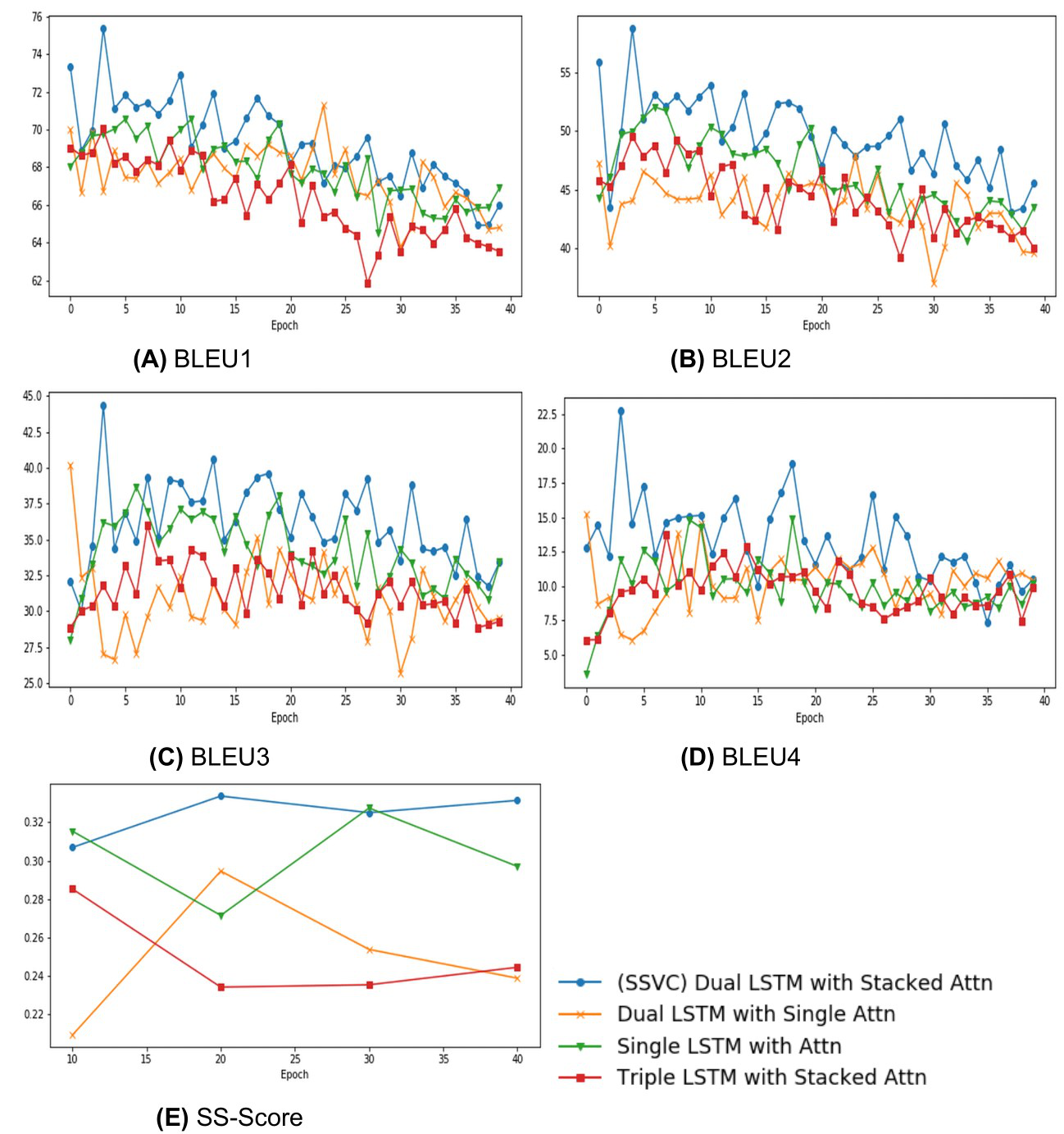}
    %\\caption:“Potatoes are being chopped by a knife”
    }
    \caption{Evaluating model performance with varied hard-pull units}
    \label{fig_join_result}
\end{figure}

To boost the captioning capability, some state-of-the-art works like \cite{pan2017video} emphasized on the importance of retrieving additional visual information. We implemented the same fundamental idea in our model with the Spatial Hard Pull. To depict the effectiveness of our Spatial Hard Pull (SHP), we conducted experiments with our stacked attention as a constant and changed the SHP FC units with 0, 45 and 60 units successively. Figure ~\ref{fig_join_result} shows that as the number of SHP FC units are increased from 0 to 45, both BLEU and SS Score get better and again gradually falls from 45 to 60. The performance improvement in the early stages indicate that SHP layer is indeed improving the model. The reason for fall of scores in the later stages is that the model starts to show high variance. Hence it is evident from this analysis that our approach of using SHP layer yields satisfactory result compared to not using any SHP layer. 

\section*{Discussion}
By performing various trials on a fixed experimental setting, we analysed the spatio-temporal behaviour of a video captioning model. After seeing that single layer encoder LSTM causes more repetitive predictions, we used double and triple layer LSTM encoder to encode the visual information into a better sequence. Hence, we were able to propose our novel stacked attention mechanism with double encoder layer that performs the best among all the variations of LSTM layers that we tried. The intuition behind this mechanism is that, as our model separately gives attention to each encoder layer, this generates a better overall temporal context for it to decode the video sequence and decide whether to give more priority to the object or the action. And the addition of Spatial Hard Pull to this model bolsters its ability to identify and map high level semantic visual information. Moreover, the results also indicate that addition of excess SHP units drastically affect the performance of the model. Hence, a balance is to be maintained while increasing the SHP units so that the model does not over-fit. As a result, both of these key components of our novel model greatly contributed to improving the overall final performance of our novel architecture, that is based upon the existing fundamental concepts of state of art models.

Although the model performed good in qualitative and quantitative analysis, our proposed SS Scoring method provides greater insight to analyse video captioning models. The auto metrics although useful, cannot interpret the videos correctly. In our experimental results, we can see a steep rise in the BLEU Score in Figure ~\ref{fig_stacked_result} and Figure ~\ref{fig_join_result} at early epochs even though the predicted captions are not up to the mark. These suggest the limitations of BLEU score in judging the captions properly with a meaningful score. SS Score considers these limitations and finds a good semantic relationship between the context of the videos and the generated language that portrays the video interpreting capability of a model into language to its truest sense. Hence, we can safely evaluate the captioning capability of our Stacked Attention with Spatial Hard Pull mechanism to better understand the acceptability of the performance of our novel model.


\section*{Conclusion and Future Work}
Video captioning is a complex task. This paper shows how stacking the attention layer for a multi-layer encoder makes a more semantically accurate context. Complementing it, the Sparse Sequential Join, introduced in this paper, is able to capture the higher level features with greater efficiency.

Due to our computational limitations, our experiments use custom pre-processing and constrained training environment. We also use a single feature as input unlike most of the state-of-the-art models. Therefore, the scores we obtained in our experiments are not comparable to global benchmarks. In future, we hope to perform similar experiments with industry standard pre-processing with multiple features as input.

The paper also introduces the novel SS Score. This deterministic scoring metric has shown great promise in calculating the semantic sensibility of a generated video-caption. However, since it is a human evaluation metric, it relies heavily on human understanding. Thus, a lot of manual work is to be put behind it. For the grammar score, we can use \cite{naber2003rule}'s ``A Rule-Based Style and Grammar Checker" technique. This will partially automate the SS Scoring method.

\

\bibliography{main.bib}

\end{document}